\documentclass[11pt]{article}

\usepackage[final]{acl}

\usepackage{times}
\usepackage{latexsym}

\usepackage[T1]{fontenc}

\usepackage[utf8]{inputenc}

\usepackage{fvextra} 
\usepackage{tikz}
\usetikzlibrary{arrows.meta,positioning,fit,calc}

\usepackage{microtype}

\usepackage{inconsolata}

\usepackage{graphicx}

\usepackage{pgf-pie}
\usepackage{xcolor}

\definecolor{myred}{HTML}{FF9F9B}
\definecolor{myblue}{HTML}{A1C9F4}
\definecolor{mygreen}{HTML}{8DE5A1}
\definecolor{mygray}{HTML}{C4C2C2}

\author{
    \textbf{Nikita Borovkov}\thanks{These authors contributed equally to this work.}, \textbf{Elisei Rykov}\footnotemark[1], \textbf{Olga Tsymboi}\footnotemark[1], \textbf{Sergei Filimonov}\footnotemark[1],  
    \textbf{Nikita Surnachev}\footnotemark[1], \\ 
    \textbf{Dmitry Bitman}, \textbf{Anatolii Potapov} \\ 
    T-Tech \\
 \small{
   \textbf{Correspondence:} \href{mailto:surnachev.nikita@gmail.com}{surnachev.nikita@gmail.com}
 }
}

\title{Learning Selective LLM Autonomy from Copilot Feedback in Enterprise Customer Support Workflows}

\usepackage{hyperref} 
\usepackage{booktabs}
\usepackage{amsmath}
\usepackage{graphicx}
\usepackage{tabularx}
\newcolumntype{Y}{>{\raggedright\arraybackslash}X}
\usepackage{multirow}
\usepackage{float}

\usepackage{tikz}
\usepackage{pgf-pie}

\begin{document}
\maketitle

\begin{abstract}
We present a deployed system that automates
end-to-end customer support workflows inside
an enterprise Business Process Management
(BPM) platform. The approach is scalable in
production and reaches selective automation within two weeks for a new process, leveraging supervision already generated at scale: structured per-case UI interaction traces and low-overhead copilot feedback,
where operators either accept a suggestion or provide a correction. A staged deployment pipeline
trains a next UI action policy, learns a critic from
copilot feedback to calibrate abstention, and executes only high-confidence steps in the background while deferring uncertain decisions to operators and resuming from the updated UI state.
This setup lets one operator supervise multiple
concurrent sessions and be interrupted only when the system is uncertain. The system operates on a schema-driven view of the BPM interface and
includes monitoring and safe fallbacks for production. In production, it automated 45\% of
sessions and reduced average handling time by
39\% without degrading support quality level.
\end{abstract}

\section{Introduction}
LLM-based agents promise end-to-end automation by reasoning and acting with tools \citep{yao2023react,schick2023toolformer} and show strong results on web and GUI interaction benchmarks \citep{osworld,drouin2024workarena}. However, deploying such agents in enterprise customer support remains hard: they interleave many model calls with UI actions, so small errors propagate over long horizons; multi-step control loops are costly at scale, and frequent UI and process changes require conservative uncertainty handling to avoid quality regressions. Traditional automation avoids some of these risks but is expensive to build and maintain, relying on manually authored dialogue flows \citep{qin2023etodSurvey} or brittle Robotic Process Automation~(RPA) scripts \citep{hindel2020rpa}.

We present a deployed alternative for an enterprise BPM system: instead of hand-authoring automation, we learn workflows directly from operator behavior, reaching selective automation within two weeks without manual rule authoring. Our setting provides two scalable supervision signals: each case yields structured UI interaction traces, and copilot mode provides lightweight feedback: operators either accept a suggestion or correct it.

Using a staged deployment procedure inspired by imitation learning \citep{dagger,aggrevate}, we progress from offline training on traces to copilot supervision and then to gated background execution, where only high-confidence actions are automated. The system is fully deployed and serves production traffic; we report online A/B results on approximately 15k customers per group, demonstrating a 39\% reduction in operator active time without degrading support quality.

Our system combines an LLM policy trained on serialized state--action traces, a critic trained on trajectory feedback labels to support abstention \citep{selectivenet,hendrickx2024rejectSurvey}, and a controller that executes only above a calibrated confidence threshold, otherwise deferring to the operator and resuming from the updated UI state. To improve robustness under UI change, the system operates on a schema-driven JSON representation of the BPM interface rather than screenshots.

Our main contribution is a production-scalable staged deployment process that turns BPM UI traces and lightweight copilot feedback into safe, gated background execution. We also report the key engineering components—schema-driven state/action serialization, deployment safeguards (abstention, data drift handling, fallbacks, monitoring), and an online evaluation protocol with ablations—together with measured gains in operator active time.

\section{Related Work}
Customer support automation has long relied on deployed dialogue and agent-assist systems that suggest replies, retrieve internal knowledge, or summarize conversations for operators \cite{fadnis-etal-2020-agent,obadinma-etal-2022-bringing,feigenblat-etal-2021-tweetsumm-dialog}. Newer datasets also capture customer-support interactions, from transactional dialogues to more natural service conversations \cite{byrne-etal-2021-tickettalk,gung-etal-2023-natcs}. Most prior work targets conversation-level assistance or knowledge access, while our setting requires safe execution of multi-step UI procedures inside an enterprise BPM interface.

Early approaches framed UI and workflow control as behavior cloning from interaction logs \cite{alvinn}. Over long horizons, covariate shift makes errors compound, which motivates interactive imitation learning methods such as DAgger and AggreVaTe that gather expert supervision on states reached by the learned policy \cite{dagger,aggrevate,aggrevated}.

GUI agents also differ in what they observe. Text-based methods act on structured interface signals such as HTML, DOM, or accessibility trees \cite{deng2023mind2web,gur2024realworldwebagentplanninglong,lai2024autowebglmlargelanguagemodelbased}, while multimodal agents additionally use screen information \cite{lu2024omniparserpurevisionbased,Xu_Chen_Wang_Liu_2025,tan2024cradleempoweringfoundationagents}. In deployed systems, raw DOM is often verbose and fragile under UI changes, which motivates intermediate representations and stronger action grounding \cite{wu2024osatlasfoundationactionmodel,gou2025navigatingdigitalworldhumans}. We use the more controlled nature of enterprise BPM to extract a schema-driven JSON state from the DOM that keeps only task-relevant entities and actionable controls, enabling monitoring and reliable tool calling without screenshots. This aligns with industry efforts that restructure support policies and workflows into LLM-friendly formats \cite{su-etal-2025-llm}.

Our deployment loop is DAgger-like. Operator rejections trigger corrective supervision in the same UI state, and we execute selectively with low-confidence handoff to reduce drift over long procedures. Instead of preference-based alignment methods, we use lightweight operator feedback to train a critic and manage the error–coverage trade-off needed for safe background automation \cite{ppo,rafailov2023dpo,grpo}.

\section{Framework}

\begin{figure*}[!htpb]
  \centering
  \includegraphics[width=0.8\textwidth]{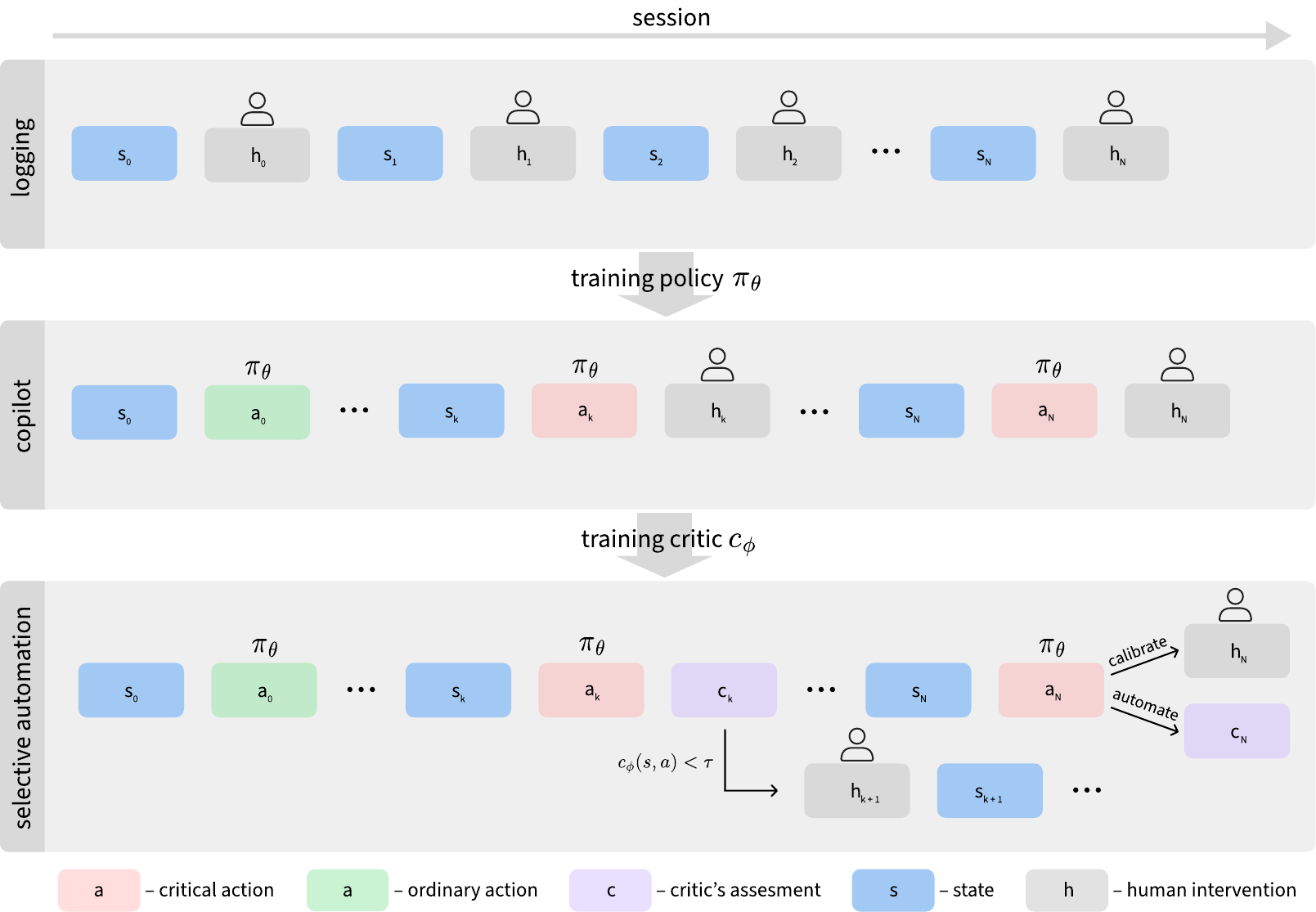}
  \caption{\textbf{Staged deployment for selective automation.}
\emph{Logging} collects operator trajectories with UI context to train an initial policy.
\emph{Copilot} deploys the policy to propose actions with human oversight on critical steps, producing accept or override feedback.
\emph{Selective automation} adds a critic that scores policy proposals and executes only critic-approved critical actions.
This stage runs in calibration with mandatory human review at the finalizing step and transitions to automation where sessions can be executed end-to-end under calibrated threshold(s) $\tau$.}

  \label{fig:draft-stages}
\end{figure*}

Our approach involves a staged deployment algorithm, progressing through three distinct phases: logging, copilot, and selective automation, each building on the data gathered in the previous stage, allowing for gradual integration into the production environment. The deployment is designed to collect data continuously at each stage, ensuring that each transition is based on robust operational feedback.
\paragraph{Problem Setting}
Operators resolve customer issues via a BPM system integrating full-history chat, scenario search, a workflow workspace for guided UI interactions (e.g., credit requests, account checks, data updates), and auxiliary tabs showing customer products and statuses. Multiple scenarios can run concurrently, with operators interleaving structured UI actions and free-form chat, relying on both encoded business rules and tacit knowledge.

A support session is a trajectory of states $s_t$ and actions $a_t$: $\{s_t, a_t\}_{t=1}^T$,
where $s_t$ comprises chat, UI, scenario, and auxiliary context, and $a_t$ spans UI navigation and interaction, tool invocation, chat messages, or session closure. The automation objective is selective: maximize automated step coverage while deferring uncertain actions to humans to maintain correctness and customer experience.

\paragraph{Data}
Operator sessions are fully logged, capturing UI state, events such as clicks, edits, submissions, and procedure openings, chat streams, and metadata including timestamps, business line, and routing details. These traces are serialized into structured state-action pairs for behavior cloning. State serialization produces JSON snapshots of the last 30 chat messages, relevant UI context such as customer profile, active scenario, and global announcements, along with operator actions. Actions fall into four categories~--- message composition, navigation, selection and form-filling, and session management~--- each targeting a specific interface element identified by its unique \texttt{control\_id} (see Table~\ref{tab:action_categories} in Appendix~\ref{sec:data-structure} for the full action taxonomy). We then allow sufficient time, typically a week or at least one business cycle, to gather comprehensive data reflecting operational patterns.

For LLM input, UI snapshots and actions are merged into a chronological dialog interleaving customer messages, operator actions, and updated UI states. The model predicts the next action, treating snapshots purely as contextual evidence.
Figure~\ref{fig:pipeline} summarizes our end-to-end data processing pipeline 
further details are provided in Appendix~\ref{sec:data-structure}.

\paragraph{Copilot stage}
Once the logging stage is complete, we proceed to create our first policy, $\pi_\theta(a \mid s)$, which predicts the next action $a$ based on a serialized state $s$. The policy is trained in two steps: (1) optional domain-adaptive pretraining on internal texts like knowledge bases, UI descriptions, and chatbot logs, which may boost model performance if available, and (2) direct supervised fine-tuning (SFT) on logs, requiring careful balancing and regularization. The trained models are evaluated using offline metrics, primarily action accuracy, measuring how well predicted actions match actual ones, including their arguments. Once trained, the policy is deployed in copilot mode, where the model proposes actions, and operators either accept or manually correct the suggestion, producing labeled examples of actions. If rejected, the correct markup for the step is obtained for future model iterations. In copilot mode, the system's performance is monitored through the acceptance rate, the fraction of actions accepted by operators, which is crucial in addressing the multiple paths problem. This allows us to gather human feedback on policy predictions and assess model quality. Most operator actions involve non-critical steps (like actions that do not change the BPM user state) that do not affect the client or organization, and critical actions (sending messages or terminating interaction) that do. To speed up the process, non-critical actions can be performed automatically by the model, requiring human approval only for critical ones. If an operator disapproves a critical action, they can interact with the interface until the next critical action, after which control returns to the model. This approach results in copilot logs containing accept/reject feedback only for critical actions.

\paragraph{Selective automation stage}
At this stage, we train a critic $c_\phi(s,a)\in[0,1]$ from copilot logs to estimate whether a \emph{critical} action $a$ is correct in state $s$ (see Appendix~\ref{sec:app_exs} for details on the critic  training). We split selective automation into two substages: \emph{calibration} and \emph{automation}. In calibration, the system executes critic-approved actions during the session, but the finalizing action is always reviewed by a human. The operator either approves finalization or rejects it and performs corrective steps to fix earlier mistakes or complete missed work before closing. In automation, critic-approved critical actions are executed end-to-end, while finalizing actions can remain human-gated until termination reliability targets are met.

Execution is regulated by threshold(s) $\tau$, optionally stratified by action importance. The policy proposes $\hat a\sim\pi_\theta(\cdot\mid s)$, the critic computes $c_\phi(s,\hat a)$, and the action is executed only when $c_\phi(s,\hat a)\ge\tau$; otherwise control is handed to a human. The critic is evaluated only for critical actions, while non-critical ones may be executed without this gating.

We calibrate $\tau$ to meet business precision requirements on critical actions. Among critic-approved actions, the fraction that operators would accept must exceed a target, for example $0.9$, estimated from copilot feedback. We initialize $\tau$ offline on held-out copilot data, then refine it online in the calibration substage without sweeping many thresholds. Online calibration uses a rolling estimate of critical-action precision together with session-level safety signals from calibration, such as the finalization rejection rate and the rate of corrective intervention.

\paragraph{Adaptability and Safety}
Production deployment must handle out-of-distribution inputs such as new UI elements, rare actions, and unexpected requests. To ensure robustness in production, we maintain a continuous improvement loop that keeps the system adaptable to evolving customer requests and changing operator workflows. Instead of static datasets, we use dynamically refreshed ones which are relevant for recent business cycles, leveraging built-in ground-truth signals from operator actions and feedback.

Production traffic is segmented into issue-specific clusters, allowing us to train dedicated policy–critic models and localize drift.

We continuously monitor online metrics and enforce slice-level guardrails combining model calibration and business KPIs; when triggered, automation is disabled for the affected slice and the system falls back to copilot mode to gather new supervision. Models are then retrained on a rolling window and redeployed after recalibration.

Training data is anonymized through entity masking, session-level distinctiveness, and selective field removal. This setup enables fast iterations, high adaptability to process changes and personal data security (details in Appendix~\ref{sec:adaptability_and_safety}).

\section{Experiments}

\subsection{Offline Experiments}

\paragraph{Experimental Setup}
All experiments were conducted with the Qwen2.5-Instruct-7B~\cite{qwen2.5} model on a single 8$\times$H100 node. We use the 7B size due to production inference constraints: it provides the best trade-off between latency and throughput under our serving budget, while keeping GPU memory usage within limits compared to larger checkpoints.
The full set of training hyperparameters is reported in Table~\ref{tab:hyperparameters}. 
We evaluate two models offline on historical interaction logs: an action model and a critic model. 
The action model is assessed on tool and argument selection using strict string matching for most actions, and a Levenshtein-based fuzzy matcher for text-generation actions, with normalized scores thresholded to tolerate minor paraphrases.
The evaluation set is drawn from real user interactions across two sources: copilot mode and fully automated mode. 
We report results separately, as differences in operator behavior, action timing, and available context introduce a mild distribution shift.

\paragraph{Metrics}
For the action model, we report tool- and action-level accuracy (overall and stratified by action type). The critic model is a binary classifier; we therefore report precision, recall, and F1.

\paragraph{Baseline}
As a baseline, we use prompting in the same setup but without state-action SFT and without domain pretraining. We do not include a comparison with agentic systems, as building a production-grade enterprise agent would require encoding the full business process into prompts and agent skills, which in our setting entails an enormous engineering effort comparing to training on real operator data.

\paragraph{Do we need SFT policy training?}
We evaluate several zero-shot prompted models — including leading open-source and proprietary ones — against our 7B SFT policy, all receiving identical session context (prompt in Appendix~\ref{app:prompt}). As shown in Table~\ref{tab:offline-baseline}, even the strongest prompted model reaches only 47.78\% tool accuracy compared to 79.15\% for SFT, and the gap persists across all model families and scales. The comparison does not claim general model superiority; it establishes that the task domain is sufficiently specialized that domain-specific SFT is a prerequisite for reliable BPM automation. A single prompt, even when augmented with examples, cannot match SFT performance because its limited capacity is insufficient to capture the complexity of domain-specific business logic.

\begin{table}[!htbp]
\resizebox{\linewidth}{!}{%
\begin{tabular}{@{}lcc@{}}
\toprule
\textbf{Model} & \textbf{Tool Acc.} & \textbf{Action Acc.} \\
\midrule
Qwen2.5-Instruct-7B (baseline) & 34.95 & 10.27 \\
DeepSeek V3.1~\cite{deepseekai2025deepseekv31modelcard} & 42.40 & 27.83 \\
GPT-5.2~\cite{openai2025gpt52systemcard} & 43.01 & 27.02 \\
Gemini 3 Pro~\cite{deepmind2025gemini3promodelcard} & 47.78 & 32.93 \\
Qwen3-235B-A22B~\cite{qwen3} & 44.33 & 26.18 \\
\midrule
\textbf{Ours} & \textbf{79.15} & \textbf{65.48} \\
\bottomrule
\end{tabular}
}
\caption{Action accuracy of zero-shot prompting across model scales vs. domain-specific SFT, demonstrating that task-specific fine-tuning is necessary regardless of model capacity.}
\label{tab:offline-baseline}
\end{table}

\paragraph{How should historical production logs be mixed with human rejections?}
We examine whether adding operator corrections from copilot deployment improves beyond SFT on historical logs alone. Rejected samples~--- where operators override policy suggestions and log corrective actions~--- are collected over one week of production copilot usage and are mixed with historical logs at varying ratios. A 2.5k-sample copilot-rejection test set evaluates generalization to hard cases. As shown in Table~\ref{tab:ablation-study-rejected}, adding rejected correction samples boosts accuracy on the copilot rejection set, and the best trade off is achieved by combining 170k predefined log samples with 40k rejected samples while keeping predefined accuracy nearly unchanged. 
\begin{table}[!htbp]
\small
\centering
\begin{tabular}{@{}cccc@{}}
\toprule
\multicolumn{1}{l}{\textbf{$N_{\text{predefined}}$}} & \multicolumn{1}{l}{\textbf{$N_{\text{rejected}}$}} & \multicolumn{1}{l}{\textbf{Acc$_{\text{predefined}}$}} & \multicolumn{1}{l}{\textbf{Acc$_{\text{rejected}}$}} \\ \midrule
170k                                  & 0                                     & 71.25                                                   & 14.88                                             \\
100k                                  & 10k                                 & 59.10                                                   & 17.60                                             \\
100k                                  & 20k                                 & 69.77                                                   & 21.76                                             \\
100k                                  & 40k                                 & 69.52                                                   & 25.74                                             \\
40k                                   & 40k                                 & 59.11                                                   & 19.96                                             \\
50k                                   & 20k                                 & 67.25                                                   & 22.92                                             \\
170k                                  & 40k                                & 71.07                                                   & 25.72                                             \\ \bottomrule
\end{tabular}
\caption{Effect of mixing historical logs with copilot rejected corrections on predefined and rejected accuracy.}\label{tab:ablation-study-rejected}
\end{table}
We did not observe gains from single-step preference optimization  on the same feedback, likely due to tool dataset imbalance and noisy rejections; results are in Appendix~\ref{sec:app_exs}.

\begin{figure}[!htbp]
\centering
\resizebox{\linewidth}{!}{%
\begin{tikzpicture}
\pie[
  radius=1.8,
  text=legend,
  after number = \%,
  color={myred, myblue, mygreen, mygray},
  sum=auto
]{
  6.9/Environment limitations,
  12.7/Other,
  37.2/Acceptable but rejected,
  43.2/Model error
}
\end{tikzpicture}
}
\caption{High-level breakdown of rejected suggestions.}
\label{fig:rejection-pie}
\end{figure}
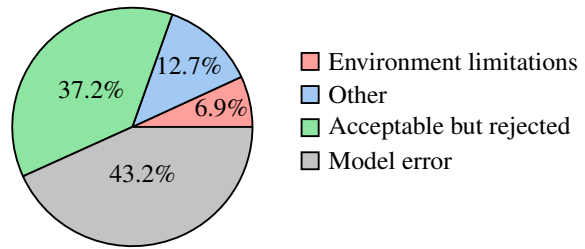

\paragraph{Why do operators reject copilot suggestions?}
We analyzed 9370 rejected actions from a 4-day period in production copilot mode to build an error taxonomy. Rejections were sorted into broad groups by action type, and human annotators reviewed 100 samples per group to find recurring patterns (see Appendix~\ref{sec:app_err_tax} for the annotation protocol and detailed category descriptions). These patterns formed a taxonomy, and proportions were estimated for all rejections. 
The results revealed four main rejection reasons, as shown in Figure~\ref{fig:rejection-pie}. 
Notably, 37.2\% of rejections came from operator preferences or optional steps, meaning the accept/reject signal is noisy and suggesting the need for more detailed feedback like graded labels. These findings highlight the importance of separating operator preferences from incorrect suggestions, better handling of environment limitations, targeted improvements for model errors in routing and context handling, as well as stronger operator onboarding around copilot usage to improve critic training and automation.

\paragraph{Additional Experiments.} We conduct ablation studies to inform key design choices in Appendix~\ref{sec:app_exs}. Scaling SFT data consistently improves accuracy, but balancing by screen or tool name offers no benefit over natural-distribution~--- uniform tool balancing actually hurts performance, suggesting real operator frequency patterns carry useful signal (Table~\ref{tab:sample-size-balancing-ablation-study}).

We use continual pretraining on anonymized operator demonstrations and QA-style synthetic data from the knowledge base~\cite{stoianov-etal-2026-pro}, which gives 5\% uplift on action accuracy.
 
Finally, an SFT-trained critic using human preference feedback outperforms a confidence-based baseline by +3.84\% precision and +3.03\% F1 (Table~\ref{tab:offline-critic}), validating explicit supervision over raw model uncertainty for action gating.

\subsection{Online Experiments}
\paragraph{Experimental Setup}
After training on copilot logs, we rolled out the policy--critic pipeline to production. We then assessed its real-world behavior on live traffic and quantified the end-to-end impact via two controlled A/B tests against production baselines. More details on A/B test groups are discussed in Appendix~\ref{app:online-experiments}.

\begin{table*}[!htbp]
\centering
\small
\begin{tabular}{@{}l l r r r r r r r @{}}
\toprule
Exp & Control$\rightarrow$Treatment & $N$/grp & Days & AAT$\downarrow$ & $\Delta$AAT & $p$ & CI \\ 
\midrule
E1 & Human-only$\rightarrow$Selective (auto) & 15.5k & 7 &
227.37$\rightarrow$139.15 & $-39\%$ &
<0.05 & [-42\%\,,\,-35\%] \\
E2 & Classic bot$\rightarrow$Selective (auto) & 77k & 32 &
124.67$\rightarrow$104.72 & $-16\%$ &
<0.05 & [-20\%\,,\,-12\%] \\
E3 & Selective (confirm)$\rightarrow$Selective (auto) & 17k & 7 &
187.66$\rightarrow$150.13 & $-25\%$ &
<0.05 & [-28\%\,,\,-23\%] \\
\bottomrule
\end{tabular}
\caption{Online customer-level A/B tests.
E1 compares a human-only BPM workflow to selective automation with automation-enabled execution.
E2 compares a classic bot baseline to our copilot pipeline with automation-enabled execution.
E3 compares selective automation with mandatory final confirmation to the same selective setup with automation-enabled execution.
CI denotes the 95\% confidence interval for the $\Delta$AAT effect.}
\label{tab:online_ab}
\end{table*}
\paragraph{Metrics}

We report \textbf{Acceptance rate} for critical suggestions in copilot mode; \textbf{Automation rate}, defined as the fraction of sessions completed end-to-end without operator intervention (sessions that only wait for customer reply are counted as automated if no reply is received within a timeout, with subsequent messages treated as a new session); \textbf{Resolution correctness}, measured via blind operator evaluation on a 0--100 rubric (Appendix~\ref{sec:appendix-quality}); and \textbf{AAT}, the average agent (operator) active time per customer, where fully automated sessions contribute zero.

\paragraph{Baselines}
We compare the copilot pipeline against two production baselines: a chatbot-assisted workflow with curated intents and scripted UI steps, and the standard human-operated workflow in the BPM without copilot or automation.

\paragraph{How well are policy suggestions accepted?}
We measure the acceptance rate of critical policy suggestions in production, averaged over one week. Operators accept deterministic UI actions at a high rate: \texttt{click\_control} (86.73\%), \texttt{close\_chat} (84.81\%), and \texttt{transfer\_chat} (91.7\%)~--- indicating strong agreement with the policy’s proposed operational steps. In contrast, \texttt{send\_text\_to\_chat} is accepted less frequently (50.81\%), which is expected because free-form responses are more sensitive to tone, wording, and policy constraints. Overall, the results suggest that procedural actions are already reliable in production, while natural-language messaging remains the main bottleneck for higher automation.

\paragraph{Does it reduce operator time?}
Compared with a human-only baseline, automation-enabled selective operation reduces AAT from 227.37 s to 139.15 s, a 39\% decrease that is statistically significant at p < 0.05, as shown in E1, Table~\ref{tab:online_ab}. Resolution correctness remains on par with the baseline, showing no statistically significant difference. In automation mode, the system fully automates 45\% of sessions.

\paragraph{Does it outperform a classic bot baseline?}
Relative to a production baseline that combines a chatbot with scripted steps, our copilot pipeline reduces AAT from 124.67 s to 104.72 s, a 16\% decrease that is statistically significant at p < 0.05, as reported in E2, Table~\ref{tab:online_ab}. Resolution correctness remains unchanged. On the operational side, new procedures are rolled out on a roughly two-week deployment cycle, and ongoing effort consists primarily of monitoring and periodic continual improvement rather than the continuous scripting and brittle workflow maintenance typical of RPA systems. Further details are provided in Appendix~\ref{sec:adaptability_and_safety}.

\paragraph{Does end-to-end automation beat copilot-style operation?}
Within the selective workflow, moving from mandatory final confirmation to automation-enabled execution reduces AAT from 187.66 s to 150.13 s, a 25\% decrease that is statistically significant at p < 0.05, as shown in E3, Table~\ref{tab:online_ab}. Resolution correctness remains statistically unchanged. This result isolates the incremental benefit of removing the confirmation bottleneck: when execution can proceed automatically, operator time shifts from repeated action-level handling toward issue diagnosis and closure validation, yielding additional throughput gains.

\section{Conclusion}

We presented a deployed system that progressively moves from copilot supervision to selective automation inside an enterprise BPM platform. Instead of full autonomy, we treat automation as incremental rollout: collect lightweight accept/override feedback from operators, then gate execution so only high-confidence steps run automatically while uncertain cases defer to humans. In production, the largest gains come from automating high-frequency, low-risk procedural work while preserving safety through abstention. The approach is more adaptable than rule-based automation and more predictable than agentic control loops, leveraging naturally occurring supervision and supporting continuous adaptation via monitoring and retraining. The recipe generalizes to any repeated-action workflow in a structured interface where interaction logs are available.

\section*{Limitations}
Our results come from proprietary data within a single enterprise ecosystem, limiting reproducibility and cross-domain generalization. This setup has not yet been evaluated in environments with less standardized workflows or less structured UI and interaction data. Experimenting with non-structured setups remains an important direction for future work. Automation is deliberately restricted to low-risk actions with operator approval on higher-risk steps, so reported gains do not reflect unconstrained autonomy. Expanding reliable coverage to long-horizon and language-sensitive steps without sacrificing safety remains underexplored. The approach depends on critic thresholds calibrated from offline proxies with limited online refinement, implicitly assuming these signals stay correlated with business metrics under shifting traffic and workflows. As automation reaches rarer scenarios with sparse supervision, both policy and critic can degrade in ways that are hard to detect in aggregated metrics, motivating conservative gating, slice-based monitoring, and rapid rollback capabilities.

\section*{Ethical Statement}
Our work deploys an LLM-based automation system for customer-support workflows. The data may include sensitive information, so all training and evaluation were conducted inside a controlled enterprise environment with strict access control. We reduce privacy risk by masking direct identifiers when possible, limiting logs to what is necessary for operation and evaluation, and applying retention and auditing practices aligned with internal security policies. All examples in this paper are synthetic or carefully redacted.

To reduce harm from incorrect automation, the system uses selective automation with action-level risk tiering. Low-risk actions, such as navigation and routine interface steps, can be executed with minimal friction. High-risk actions that may affect the customer, such as sending messages or making irreversible account changes, require explicit human approval. A learned critic can abstain and defer to an operator, or fall back to a safer baseline when confidence is low or inputs are out of distribution. Operators can review, accept, or reject proposed actions, and we monitor safety and quality signals such as deferrals, rollbacks, and task outcomes. We also check performance across heterogeneous issue types and other relevant slices, and tighten automation thresholds when risks or disparities appear.

\textbf{Operator impact.} The system is designed to scale support capacity to a growing customer base rather than to reduce headcount. In practice, automation removes repetitive procedural work and shifts operator focus toward complex cases requiring judgment. To support this transition, operators undergo onboarding that frames their evolving role as AI editors who review and refine system outputs. Qualitative feedback from operators has been positive, with operators reporting reduced burden from routine tasks.
Reproducibility is limited by proprietary data, so we report aggregated results and sufficient methodological details to support assessment.

\bibliography{custom}
\clearpage
\newpage
\appendix
\section{Data Pipeline}\label{sec:data-structure}

\begin{figure*}[!htbp]
  \centering
  \includegraphics[width=\linewidth, trim=100 100 100 100, clip]{images/data_pipeline_v2.pdf}
  \caption{Data pipeline. We first log each BPM step as an HTML snapshot that includes both the customer chat and the visible UI state. We then convert the raw HTML into a task-centric JSON representation using a set of hand-written parsers. Finally, we merge the JSON state with the dialog history to produce model-ready training/inference samples.}
  \label{fig:pipeline}
\end{figure*}
\paragraph{Interaction Logging}
We set up the BPM system to record full operator sessions. Each log entry includes the current UI state visible to the operator, covering both customer-facing and back-office information, as well as UI events like button clicks, form edits and submissions, table selections, and procedure openings. It also captures the chat history, meaning messages from the customer and replies from the operator, along with extra metadata such as timestamps in both system and customer time zones, business line, and routing details. These raw interaction logs serve as the basis for our pipeline: they are later converted into state–action pairs and used to train our policy through behavior cloning.

\paragraph{State Serialization and Action Space}
A key part of our data pipeline is state serialization. We build a structured JSON snapshot of the operator's interface and the ongoing conversation using manually designed rule-based parsers. Unlike web-browsing agents that deal with unconstrained and highly variable HTML, agents in a BPM environment work with a limited and well-defined set of interface states. We take advantage of this property to reduce noise and extract only the elements needed to resolve the customer's request.

More specifically, we serialize the last 30 chat messages, updating the window when the operator scrolls, along with the UI information used during issue resolution, such as the customer profile, the active scenario or procedure, and global context like incident announcements. We also capture the operator's actions performed in the interface.

We distinguish four categories of actions, as listed in Table~\ref{tab:action_categories}: composing and sending messages to the customer, navigation actions that update the UI state, such as switching screens or opening procedures; selection and form-filling actions that modify fields on the current screen, and session-management actions that close the current interaction or transfer the chat to another operator. Each recorded UI action targets a specific interface element identified by its unique \texttt{control\_id}.

\begin{table}[!htbp]
    \centering
    \resizebox{\linewidth}{!}{%
    \begin{tabular}{ll}
        \toprule
        \textbf{Category} & \textbf{Action} \\
        \midrule
        Message Composition & \texttt{send\_text\_to\_chat} \\
        \hline
        \multirow{2}{*}{Navigation} & \texttt{open\_procedure} \\
         & \texttt{click\_control} \\
        \midrule
        \multirow{4}{*}{Selection \& Form-filling} & \texttt{select\_radio\_button} \\
         & \texttt{select\_element\_in\_combo\_box} \\
         & \texttt{select\_element\_in\_select} \\
         & \texttt{fill\_input} \\
        \midrule
        \multirow{2}{*}{Session Management} & \texttt{close\_chat} \\
         & \texttt{transfer\_chat} \\
        \bottomrule
    \end{tabular}%
    }
    \caption{Action Categories and Commands}
    \label{tab:action_categories}
\end{table}

\paragraph{Converting Serialized States to LLM Input}
The final step is to combine all serialized features from an interaction session into a single model-ready dialog in a chat format. We build the dialog by interleaving customer and operator messages with UI interactions in chronological order, meaning we represent textual messages, UI snapshots, and actions within a single session timeline. At the beginning of the session, we place the initial UI snapshot together with the chat messages observed up to that moment. Then we iteratively add each operator action followed by the corresponding updated UI snapshot.

Both actions and snapshots are encoded as textual messages. During training, the model is trained to predict the next action, while UI snapshots serve only as contextual evidence and are not treated as prediction targets.

\section{Adaptability and Safety}\label{sec:adaptability_and_safety}

\paragraph{OOD Handling}
One of the biggest challenges in production is dealing with situations that fall outside the distribution of the training logs. Examples of such situations are new UI elements or changes in the interface layout, rare action sequences or corner cases that were never logged before, and unexpected client requests or completely new process flows.

To handle these out-of-distribution (OOD) scenarios, we put several safeguards in place:
\begin{itemize}
  \item \textbf{Daily data and log validation}: New or unknown UI elements are caught by automatic validation rules. Every day we calculate the share of sessions that fail validation and record a breakdown of failure rates for each rule. If error ratios go above the thresholds, which are set based on acceptable frontend or backend update delays, an alert is sent out for further investigation.
  \item \textbf{Critic-based uncertainty detection}: Whenever the critic score $c_\phi(s,a)$ drops below a threshold $\tau$, the action is forwarded to a human operator instead of being executed. This protects clients from hasty or poorly justified model actions in real time. It also helps us find model weaknesses later, by analyzing logs of low-confidence or rejected actions.
  \item \textbf{Fallback policies}: For cases that are still not covered, we use deterministic or rule-based fallback procedures to guarantee safety and compliance. These procedures also step in to stop execution chains that have clearly gone off track — for instance, when a loop is detected or the sequence of actions no longer matches the expected process logic.
\end{itemize}

Together, these mechanisms let the system work reliably even in new or unusual situations, while keeping safety and support quality high.

\paragraph{Continuous Improvement Loop}

The method described above forms the core of our approach, but to meet the demands of production we also need to make sure it stays flexible. In practice, the topics of customer requests keep changing, and operator scripts are updated regularly. Because of this, we need a fast and reliable way to test new ideas about model input, architecture, or small pipeline features, and to validate them on up-to-date data.

To meet these needs, we maintain a continuous training and validation loop built around the following components:
\begin{itemize}
  \item \textbf{Dynamic datasets}. Our datasets are not kept fixed for a long time. Each dataset covers one business cycle or one series of experiments and is replaced afterwards. Since our process already provides ground-truth actions and accept/reject labels from human operators, we only need a reusable set of data preparation scripts and make sure they stay up to date.
  \item \textbf{Slice-aware operation.} Since production traffic is heterogeneous, we segment sessions into issue clusters (e.g., by topic) and train a dedicated policy--critic pair per cluster. This makes drift localized: regressions typically appear in a small slice and can be diagnosed and fixed without globally tightening automation. A practical constraint is the long tail of small clusters with limited gold demonstrations; in such slices critic-based abstention may be unreliable, so we keep them in copilot-only mode (no selective automation) until sufficient supervision accumulates.
  \item \textbf{Online metrics monitoring.} Besides data quality, the quality of the support itself must also remain high. For this reason, we set up alerts for drops in online metrics and monitor them on an ongoing basis. For each cluster we track a set of guardrails that combine calibration-based estimates of critical-action precision and session-level safety signals such as finalization rejection and corrective intervention rates with business guardrails such as returns and CSAT.
  \item \textbf{Slice-level guardrails and rollback.}  When any guardrail triggers, we automatically disable selective automation for the affected cluster and fall back to copilot-only, routing critical actions to operators to collect fresh supervision. Using these newly collected operator traces, we retrain the policy and critic on a rolling window and rerun the standard selective-automation pipeline, including mandatory recalibration of the critic threshold, before redeploying the updated pair in calibration and re-enabling automation.

\end{itemize}
These practices give us high adaptability to process changes and short time-to-market for new improvements.

\paragraph{Safety and Compliance Constraints}
In a real-world customer support setting, production logs inevitably contain many kinds of personal and sensitive information, such as customer names, contact details, account identifiers, and so on.

To comply with privacy regulations and internal security policies, all datasets used for model training, validation, and evaluation go through strict anonymization and filtering procedures. Before any data enters the training pipeline, we carry out the following steps:

\begin{itemize}
  \item \textbf{Data classification and masking}: Each log record is parsed to detect personal data fields using a combination of rule-based detectors (regular expressions, entity dictionaries) and trained named entity recognition (NER) models. Detected entities are replaced with neutral placeholders (e.g., \verb|<NAME>|, \verb|<PHONE_NUMBER>|), so that model inputs keep their structure and meaning but contain no personal content.
  \item \textbf{Session-level distinctiveness}: Some identifiers need to remain distinguishable within the same session. To achieve this, different personal data instances of the same type are replaced with masks that have different letter suffixes (e.g., \verb|<NAME_A>|, \verb|<PHONE_NUMBER_B>|).
  \item \textbf{Selective field removal}: For certain data types, such as attachments, contract documents, or images, the entire field is removed from the dataset to eliminate any chance of leakage.
\end{itemize}

These safeguards ensure that no personal information is used during model training or evaluation.
\section{Offline Experiments}\label{sec:app_exs}
\paragraph{Hyperparameters} Table~\ref{tab:hyperparameters} summarizes the full set of SFT training hyperparameters.
\begin{table}[!htbp]
\centering
\small
\begin{tabular}{@{}ll@{}}
\toprule
\textbf{Parameter}               & \textbf{Value}        \\ \midrule
\multicolumn{2}{l}{\textbf{Model}}                       \\
Base model                       & Qwen2.5-Instruct-7B  \\
Max sequence length              & 32768                 \\
Precision                        & bfloat16              \\ \midrule
\multicolumn{2}{l}{\textbf{Training}}                    \\
Training duration                & 2 epochs              \\
Global train batch size          & 32                    \\
Microbatch size (per device)     & 1                     \\
Gradient clipping                & norm, threshold 2.0   \\ \midrule
\multicolumn{2}{l}{\textbf{Optimizer and scheduler}}     \\
Optimizer                        & Decoupled AdamW       \\
Learning rate                    & $2 \cdot 10^{-6}$     \\
$\beta_1, \beta_2$               & (0.9, 0.95)           \\
$\epsilon$                       & $10^{-8}$             \\
Weight decay                     & $10^{-7}$             \\
LR schedule                      & cosine with warmup    \\
Warmup                           & 500 batches           \\
Final LR fraction ($\alpha_f$)   & 0.1                   \\ \midrule
\multicolumn{2}{l}{\textbf{Evaluation}}                  \\
Eval interval                    & every 500 batches     \\
Eval subset                      & 20 batches            \\ \midrule
\multicolumn{2}{l}{\textbf{Distributed training (FSDP)}} \\
Sharding strategy                & FULL\_SHARD           \\
State dict type                  & sharded               \\
Activation checkpointing         & enabled               \\
Forward prefetch                 & enabled               \\
Backward prefetch                & BACKWARD\_PRE         \\
CPU offload                      & disabled              \\ \bottomrule
\end{tabular}
\caption{Hyperparameters for SFT model trainig.}\label{tab:hyperparameters}
\end{table}
\paragraph{Sample Size and Balancing Ablation}
We study how the amount of SFT data and different balancing strategies affect policy performance. We compare three sampling strategies for training sets of size $N$ from 4k to 64k:
\begin{itemize}
\item Random sampling from the natural state-action production logs distribution without any balancing.
\item Balancing by BPM screen, where we sample uniformly across screen buckets.
\item Uniform balancing by the tool name that is referenced in the action.
\end{itemize}
All models are trained with the same architecture, hyperparameters, and we evaluate them on a held-out set using argument accuracy. The results in Table~\ref{tab:sample-size-balancing-ablation-study} show that increasing the dataset size consistently improves performance. However, we do not observe a consistent benefit from balancing compared to random sampling. In fact, balancing by tool name performs worst across all sample sizes. This finding suggests that making tool frequencies uniform removes the informative structure of real operator behavior and harms the model's ability to generalize.

\begin{table}[!htbp]
\centering
\small
\begin{tabular}{@{}cccccc@{}}
\toprule
\textbf{Dataset Size} & \textbf{No Balancing} & \textbf{By Screen} & \textbf{By Tool} \\ \midrule
4k          & \textbf{55.92}       & 55.80            & 50.97          \\
8k          & 61.47       & \textbf{62.58}            & 57.47          \\
16k         & \textbf{67.15}       & 66.82            & 62.02          \\
32k         & \textbf{69.60}       & 69.22            & 67.47          \\
64k         & \textbf{71.19}       & 70.77            & 69.28          \\ \bottomrule
\end{tabular}
\caption{Argument accuracy on the held-out set for different training set sizes and balancing strategies. Increasing the dataset size consistently improves performance, while balancing by tool name leads to the lowest accuracy across all sizes. Best results per row are highlighted in bold.}\label{tab:sample-size-balancing-ablation-study}
\end{table}

\paragraph{Critic Architecture and Training}

The critic $c_{\phi}(s,a)$ is initialized from the same Qwen2.5-7B checkpoint as the policy and fine-tuned as a binary classifier on copilot accept/reject labels. The input is the same serialized state–action dialog used by the policy, appended with the proposed action; the model is trained to generate a single token (accept or reject) with the softmax probability of accept serving as the scalar score $c_{\phi}(s,a) \in [0,1]$. We train the critic using the same fine-tuning hyperparameters as the policy SFT stage. We found this generative framing more stable than adding a classification head, as it reuses the policy's learned representations without introducing new parameters. The critic is evaluated only on critical actions at inference time; non-critical actions bypass the gate entirely.

\paragraph{Comparing uncertainty-based critic with SFT.}
\begin{table}[!htbp]
\small
\centering
\begin{tabular}{@{}llll@{}}
\toprule
\textbf{Method} & \textbf{Precision} & \textbf{Recall} & \textbf{F1} \\ \midrule
Action confidence & 70.84              & 73.43           & 70.96       \\
SFT             & \textbf{74.68}              & \textbf{73.53}           & \textbf{73.99}       \\ \bottomrule
\end{tabular}
\caption{Offline evaluation of critic model variants on binary action decision prediction, where the model decides whether to execute or defer a proposed action. The SFT critic yields notable gains in Precision and F1 while maintaining comparable recall.}\label{tab:offline-critic}
\end{table}
We evaluate two approaches for the critic model. The task is framed as a binary classification problem: the model predicts whether a proposed action should be executed or rejected. As a baseline, we use the action model's overall confidence score as a proxy for executability. In other words, we treat the model's confidence in the predicted action as the critic signal. Our main approach trains a dedicated critic through SFT. This critic is initialized from the action model and then fine-tuned on historical interaction logs that contain human feedback on suggested actions, specifically whether they were accepted or rejected. We report Precision, Recall, and F1 on an offline test set.

The results in Table~\ref{tab:offline-critic} show that the critic trained with SFT consistently outperforms the confidence-based baseline. Precision improves by 3.84\%, and F1 increases by 3.03\%, while recall remains at a comparable level with only a 0.10 \% difference. These findings suggest that learning from historical accept and reject feedback produces a more reliable execution signal than relying on the action model's raw confidence alone.

\paragraph{Offline Single-Step Preference Tuning}

\label{app:pref-tuning}

Learning from operator feedback is important for \emph{selective automation}: incorporating corrections on the cases where the copilot policy fails increases coverage while maintaining safety. In our main approach, we already leverage this feedback via SFT by training on copilot-mode failures, replacing the rejected policy action with the operator-corrected action.

However, human feedback is often used with \emph{preference-based} alignment objectives rather than pure supervised learning. We therefore evaluate offline single-step preference tuning on copilot data by constructing paired examples from operator overrides: for a rejected proposal $a^{-}$ in state $s$, we treat the operator correction $a^{+}$ as preferred and optimize Direct Preference Optimization (DPO, \cite{rafailov2023direct}) and Anchored Preference Optimization (APO, \cite{d-oosterlinck-etal-2025-anchored}) on $(s, a^{+} \succ a^{-})$ pairs.

Copilot feedback can only be used in a \emph{single-step} setup in our logging regime: when an operator rejects an action, the policy’s action chain is interrupted and we do not observe the continuation of the model’s trajectory, preventing multi-turn preference learning from these logs alone.

Empirically, single-step preference tuning degrades performance on the production distribution (Table~\ref{tab:single-turn-preference}). A likely cause is \emph{distribution shift}: constructing a paired dataset from rejected events over-weights difficult, operator-intervened states and moves the training distribution away from historical production logs, which hurts metrics on the production bucket.

In addition, operator rejections are noisy as a preference signal. Figure~\ref{fig:rejection-pie} shows that a large fraction of rejected actions were still \emph{acceptable}, implying that many $(a^{+} \succ a^{-})$ pairs encode stylistic or risk-aversion preferences rather than correctness, injecting substantial label noise into the preference dataset.

Finally, copilot rejection pairs introduce a strong \emph{tool imbalance} and repeated “template” pairs (e.g., the operator selecting \texttt{click\_control} instead of \texttt{send\_text\_to\_chat} in similar contexts). This makes it easy for the model to \emph{hack} the objective by systematically shifting probability mass toward a frequent tool, rather than learning the causal decision rules needed for correct action selection.

A promising direction is multi-turn alignment, which would address most of the above issues: operators can more reliably judge full action sequences than isolated steps under path multiplicity (an individual step may be acceptable depending on what the policy does next); and trajectory-level feedback reduces action/tool imbalance since supervision is assigned to an entire rollout rather than a single tool choice. However, moving to multi-turn alignment is technically challenging: it requires an offline environment that enables policy exploration in BPM engine and assigns outcome-based rewards to completed trajectories. We leave this direction for future work.

\begin{table}[!htbp]
\small
\centering
\begin{tabular}{@{}lc@{}}
\toprule
\textbf{Method} & \textbf{Action Acc. (\%)} \\ \midrule
SFT (baseline) & \textbf{76.03} \\
SFT + APO & 68.73 \scriptsize{\textcolor{red}{(-7.30)}} \\
SFT + DPO & 69.97 \scriptsize{\textcolor{red}{(-6.06)}} \\ \bottomrule
\end{tabular}
\caption{Effect of single-step preference tuning on action prediction accuracy. Both DPO and APO models are initialized from the SFT baseline.}\label{tab:single-turn-preference}
\end{table}

\section{Online Experiments}\label{app:online-experiments}
This appendix provides additional information on online experiments and A/B tests that have been performed.

All tests use customer-level randomization, so control and test are exposed to the same seasonality and campaigns, and receive comparable mixes of customer issues. Escalations are handled by the same pool of operators in both groups under identical staffing and SLAs. Although randomization is at the customer level, we verified balance by comparing pre-period (pre-experiment) metrics between groups, including traffic volume, escalation rate, and baseline AAT, and found no material differences.

We compare against two established production baselines. Human-only is the standard workflow where customer chats are handled end-to-end by human operators without copilot or automation. Classic bot is a production rule-based chatbot that routes users into curated intent flows for common issues. When the bot cannot resolve the request---e.g., the user presses the ``issue not resolved'' button or explicitly asks to talk to an operator---it hands off the chat to a human operator, who continues in the same workflow.

\section{Error Taxonomy of Rejected Copilot Suggestions}\label{sec:app_err_tax}

\begin{table*}[t]
\centering
\small
\resizebox{\linewidth}{!}{%
\begin{tabular}{@{}lrr@{}}
\toprule
\textbf{Bucket (tool-oriented)} & \textbf{\#} & \textbf{\%} \\
\midrule
C1: model suggests a templated/text action, operator performs a different BPM action & 2065 & 22.0 \\
C2: model suggests free-form text, operator performs a different non-text action & 1879 & 20.1 \\
C3: model suggests closing, operator continues the workflow & 1899 & 20.3 \\
C4: model and operator both send text but content differs & 1792 & 19.1 \\
C5: operator edits/softens the suggested text & 766 & 8.2 \\
C6: operator performs a different UI action/control within a procedure & 606 & 6.5 \\
C7: model suggests a UI click, operator writes in chat instead & 363 & 3.9 \\
\midrule
\textbf{Total} & 9370 & 100.0 \\
\bottomrule
\end{tabular}%
}
\caption{Intermediate tool-oriented grouping of rejected events (counts and percentages over 9370 rejections).}
\label{tab:rejection-buckets}
\end{table*}

\subsection{Annotation protocol.}
We start by sorting all rejected events into broad groups based on the type of action the system suggested and what the operator actually did instead. For example, the policy might suggest sending a text message to the chat, but the operator rejects it and performs a completely different action. We use simple rules to create these groups. Table~\ref{tab:rejection-buckets} shows the seven groups we used during the analysis. These groups capture how rejections look at the level of specific tools or actions, before we merge them into broader categories of causes.

For each group, human annotators look at 100 randomly selected examples. They review the full context, including the serialized state, the suggestion made by the policy, and the action the operator chose instead. Annotators write free-form notes explaining why the rejection happened and assign each example an error category and a more specific subcategory. For instance, an annotator might note that the model action is actually acceptable, with the subcategory being that the operator simply added softening language or emojis to the message. We then combine all observed categories into a single unified taxonomy and estimate how common each category is by scaling the proportions from the annotated samples to the full set of rejections collected over four days. It is worth noting that the subcategory proportions are approximate, since they come from samples of 100 examples per group and are then projected onto the entire dataset. Figure~\ref{fig:rejection-pie} shows the estimated distribution of rejection reasons across four high-level categories.

\subsection{Typical categories and failure modes.}
We further summarize the most common patterns observed within the four high-level categories.

\paragraph{Acceptable but rejected (37.2\%).}
These cases reflect divergence between the model's suggestion and operator preference or workflow style rather than a strictly incorrect action:
\begin{itemize}
    \item \textbf{Minor text edits:} the operator rewrites the suggested message to be more polite, softer, shorter, or to add small clarifying details.
    \item \textbf{Optional verification steps:} the operator performs a non-mandatory check (e.g., opening an additional tab/panel) before sending a message or closing.
    \item \textbf{Valid alternative workflow variants:} the operator follows a different but permissible procedure variant (e.g., performs an intermediate step before a standard action).
    \item \textbf{Preference-driven handling of edge cases:} e.g., foreign-language chats where operators prefer to rewrite the message manually even if the model output is reasonable.
\end{itemize}
\noindent\textbf{Implication:}
A simple binary choice between accept and reject mixes up personal preference with actual correctness. A practical next step would be to collect more detailed, graded feedback. For example, annotators could choose from options like acceptable as-is, acceptable with minor edits, incorrect, missing verification, or wrong workflow route. This would help separate the training signals that are truly useful from those that just reflect individual style.

\paragraph{Environment limitations (6.9\%).}
Here the rejection is driven by missing capabilities or incomplete observability:
\begin{itemize}
    \item \textbf{Non-text modalities:} customers provide key information via images attachments that are not available to a text-only model.
    \item \textbf{Unsupported actions:} workflows requiring phone calls or call outcomes that the copilot is not allowed or able to execute.
    \item \textbf{Logging and UI observability gaps:} in some cases, important data is missing or unclear due to issues with logging, such as certain forms not being tracked, forwarded messages not being captured, or bugs in the interface. This prevents the model from seeing the evidence it needs.
    \item \textbf{External knowledge not present in the environment:} operators sometimes use information that exists outside the platform, for example details about ongoing operational incidents, which is not shown in the interface or included in the context.
\end{itemize}
\noindent\textbf{Implication:} These cases define the boundary of safe automation and motivate explicit stop-lists, fallbacks, and future modality expansion.

\paragraph{Model error (43.2\%).}
These are cases where the suggestion is incorrect given the available context and supported capabilities. The most frequent failure modes include:
\begin{itemize}
    \item \textbf{Workflow routing and navigation errors:} the policy navigates to the wrong screen or procedure, selects a wrong control, or fails to switch between relevant UI panels. This aligns with our oracle plan experiment, where providing a gold screen plan primarily improves routing-related tools.
    \item \textbf{Missing required steps or premature termination:} the policy suggests closing the conversation or sending a message when the correct resolution requires additional actions, or a chain of actions, to be completed first.
    \item \textbf{Conversation understanding failures:} ignoring the latest user question, answering only one of multiple questions, or producing a response that does not address the user intent.
    \item \textbf{Text generation quality issues:} repeated content, overly confident claims that are not supported by evidence, and time-sensitive phrasing mistakes such as inappropriate greeting behaviour.
    \item \textbf{Context construction and truncation artifacts:} manual inspection suggests that a significant share of errors may be caused by the way context is assembled and trimmed. For example, crucial messages or action history may be missing from the input. In our internal analysis, such artifacts could account for up to roughly 26\% of rejection cases, which motivates systematic context logging, validation, and more robust trimming strategies.
    \item \textbf{Template and standard-form mismatches:} cases where the model should send, or should not send, a standard-form or template message, or where it should verify user or account state before sending. Internal estimates suggest that up to roughly 9\% of rejections may be related to such template usage patterns.
\end{itemize}
\noindent\textbf{Implication:} The mix of errors motivates several directions. First, selective execution and per-screen thresholds can help reduce long-horizon drift. Second, planner-like context can reduce routing ambiguity. Third, separating tool actions from free-form text generation can help address text-specific errors. Continuous monitoring targeting these failure modes should support all three directions.

\paragraph{Other (12.7\%).}
This bucket contains heterogeneous, low-frequency patterns that do not form stable subcategories in the 4-day sample. We treat it as a tail distribution and prioritize categories above for targeted data collection and mitigation.

A key observation is that rejections do not always mean the suggestion was wrong. A substantial share of them, 37.2\%, corresponds to actions that are acceptable but still rejected. This can happen because of operator preference, optional steps, or minor edits. This finding highlights that binary accept or reject feedback is inherently noisy in a copilot setting and should be treated as an imperfect supervision signal. In practice, reducing this noise requires deliberate operator training and onboarding. Operator groups need guidance on how to work with copilot suggestions, when to accept them, and how to handle minor adjustments. In addition, collecting richer feedback, such as graded labels or standardised rejection reason codes, would make the signal more consistent and actionable. This in turn would enable more reliable critic training and more effective automation improvements.

\section{Resolution Correctness Criteria}
\label{sec:appendix-quality}

This appendix presents the methodology used by the Quality Department to evaluate employee performance during customer interactions. Each interaction is assigned a resolution correctness score ranging from 0 to 100 points.  

The daily sample size was chosen to support 95\% confidence with approximately $\pm 3$ points precision assuming $\sigma=30$. While 400 chats are sampled per slice per day, statistical power improves as samples accumulate over time.
Quality was assessed using the Quality Department’s standard employee evaluation rubric for both the baseline workflow and the copilot-enabled workflow, including selective automation sessions handled fully by the copilot. This is possible because the copilot is trained via behavior cloning to follow the same operating procedures as human agents, so no copilot-specific scoring form is required. Each day, 400 chats were randomly sampled per slice scored on a 0–100 resolution correctness scale by four trained evaluators. Inter-rater agreement was monitored using a 3-way overlap design on a subset of chats. Twelve trained evaluators annotated 100 chats blindly each per day, producing 400 unique chats per day with each chat independently scored by three evaluators. We quantified consistency of the final 0--100 quality score using the intraclass correlation coefficient (ICC; absolute agreement), which indicated substantial agreement (ICC $\approx$ 0.8).

To reduce automation bias, the same standardized rubric and identical scoring criteria were applied to all interactions regardless of assistance condition. The evaluation is conducted using a structured form consisting of several blocks, which are listed in Table \ref{tab:quality-fullwidth-softskills}. Evaluators select a specific issue corresponding to each observed criterion in each block.  
Depending on the severity of that issue, points may be deducted from the total score. The first block includes general criteria that apply to the communication as a whole. The remaining blocks describe specific types of issues that may occur within the interaction. When a specific issue is identified, points are deducted from the overall score in proportion to its severity.

\begin{table*}[!htbp]
\centering
\small
\resizebox{\linewidth}{!}{%
\begin{tabular}{@{}lll@{}}
\toprule
\textbf{Block}                             & \textbf{Criterion}                               & \textbf{Options / Issues}                                                                                                                                                                                                                                                                                                                                                                                                                                                                                                                                                                     \\ \midrule
General                                    & Whether the Customer’s Issue was Resolved        & \begin{tabular}[c]{@{}l@{}}Issue resolved \\ Issue not resolved \\ Provided recommendations for resolution \\ Impossible to resolve \\ System limitations\end{tabular}                                                                                                                                                                                                                                                                                                                                                                                                                        \\ \midrule
\multirow{5}{*}{Customer Issue Resolution} & Understanding of the Customer’s Question or Need & \begin{tabular}[c]{@{}l@{}}Did not attempt to understand the issue \\ Ignored a direct question \\ Failed to identify the reason for service termination \\ Did not initiate contact with the customer \\ Provoked unnecessary questions from the customer \\ Unfamiliar with prior message history\end{tabular}                                                                                                                                                                                                                                                                              \\ \cmidrule(l){2-3} 
                                           & Correctness of Consultation                      & \begin{tabular}[c]{@{}l@{}}Provided misinformation, critical consequences \\ Provided misinformation, no critical consequences \\ Failed to communicate information, critical consequences \\ Failed to communicate information, no critical consequences \\ Performed an action not requested by the customer\end{tabular}                                                                                                                                                                                                                                                                   \\ \cmidrule(l){2-3} 
                                           & Interaction with the system                      & \begin{tabular}[c]{@{}l@{}}Entered incorrect data \\ Failed to open the required script \\ Closed the chat out of time \\ Other system error, critical consequences \\ Other system error, no critical consequences \\ Failed to complete identity verification \\ Incorrect transfer \\ Failed to escalate a case, critical consequences \\ Failed to escalate a case, no critical consequences \\ Incorrect session termination or disconnection \\ Did not offer unified customer proposition (UCP) \\ Incorrectly marked UCP status \\ Did not handle UCP-related objections\end{tabular} \\ \cmidrule(l){2-3} 
                                           & Resolution Speed                                 & \begin{tabular}[c]{@{}l@{}}Provided irrelevant information \\ Kept the customer waiting excessively\end{tabular}                                                                                                                                                                                                                                                                                                                                                                                                                                                                              \\ \cmidrule(l){2-3} 
                                           & Escalation (Requesting Assistance)               & \begin{tabular}[c]{@{}l@{}}Failed to contact the support line, critical consequences \\ Failed to contact the support line, no critical consequences \\ Did not attempt to resolve the issue independently\end{tabular}                                                                                                                                                                                                                                                                                                                                                                       \\ \midrule
Soft Skills                                & Vocabulary and Word Choice                       & \begin{tabular}[c]{@{}l@{}}Filler words \\ Overly templated or dry phrases \\ Did not greet the customer \\ Improper or impolite form of address \\ Overly long or rambling speech\end{tabular}                                                                                                                                                                                                                                                                                                                                                                                               \\ \cmidrule(l){2-3} 
                                           & Communication Manner and Tone                    & \begin{tabular}[c]{@{}l@{}}Interrupted the customer fewer than 2 times \\ Interrupted the customer more than 2 times \\ Engaged in argument \\ Unacceptable tone \\ Inappropriate or unpleasant phrasing \\ Dull or disengaged voice \\ Audible sighs\end{tabular}                                                                                                                                                                                                                                                                                                                            \\ \cmidrule(l){2-3} 
                                           & Clarity of Explanation                           & \begin{tabular}[c]{@{}l@{}}Not clear, critical consequences \\ Not clear, no critical consequences\end{tabular}                                                                                                                                                                                                                                                                                                                                                                                                                                                                               \\ \cmidrule(l){2-3} 
                                           & Grammatical Accuracy and Fluency                 & \begin{tabular}[c]{@{}l@{}}More than two grammatical errors \\ Fewer than two grammatical errors\end{tabular}                                                                                                                                                                                                                                                                                                                                                                                                                                                                                 \\ \cmidrule(l){2-3} 
                                           & Handling of Negative Customer Emotions           & \begin{tabular}[c]{@{}l@{}}Made no attempt to smooth out negativity \\ Provoked additional negativity \\ Did not apologize \\ No empathy or acknowledgment\end{tabular}                                                                                                                                                                                                                                                                                                                                                                                                                       \\ \cmidrule(l){2-3}
                                           & Other                                            & \begin{tabular}[c]{@{}l@{}}Discussed sensitive topics \\ Damaged brand image \\ Self-identification error \\ Looping, repeated words or messages\end{tabular}                                                                                                                                                                                                                                                                                                                                                                                                                                 \\ \bottomrule
\end{tabular}
}
\caption{Quality assessment form with detailed options for each criterion.}
\label{tab:quality-fullwidth-softskills}
\end{table*}

\onecolumn
\section{Zero-Shot Baseline Prompt}\label{app:prompt}
\begin{figure}[!htbp]
\centering
\begin{minipage}{0.98\linewidth}
\begin{Verbatim}[fontsize=\small, frame=single, breaklines=true, breakanywhere=true]
GENERAL INSTRUCTIONS:
You are assisting an operator of a bank's support chat. Operator will control all your steps so listen to
all messages from "system" role and clearly describe the reasoning behind your actions. The following messages from the "system"
role do not override these instructions, but only indicate how to proceed with a particular step.
You'll get messages from bank clients and you must use tool calling to solve their problems.
You must hold a conversation in Russian language.

Use information about client to call a proper tool with parameters

PROCEDURES:
Your most common task will be interacting with procedures. Procedures are our internal tool to solve client problems. They contain information for you,
messages for client, or even scripts to manage client products. Use tool to interact with procedures that are displayed on operators screen.
Your called tools will be performed on operators computer.

PROCEDURE INTERACTION RULES:
1. Search procedure as first interaction;
2. When call search procedure use one most common word.
3. Remember that its cheaper for us if you do action in procedure rather than you write message to client;
4. Don't communicate with client until you interact with procedure;
5. If there is a form to fill on the procedure screen fill it accordingly with respect to the information from the client;
6. Write message only after you have interacted with the procedure and have the solution for the clients problem;
7. Don't write message to client if there is actions to do in procedure, while procedure is still in progress;
8. Don't close chat until you solve clients problem. Closing with timer is more common than close without;
9. Don't wait for clients response until you wrote him a message;
\end{Verbatim}
\end{minipage}
\caption{Prompt template used for the prompting baseline.}
\label{fig:prompt}
\end{figure}

\end{document}